# Berry Twist: a Twisting-Tube Soft Robotic Gripper for Blackberry Harvesting

Johannes F. Elfferich, Ebrahim Shahabi, Cosimo Della Santina, and Dimitra Dodou

*Abstract*— As global demand for fruits and vegetables continues to rise, the agricultural industry faces challenges in securing adequate labor. Robotic harvesting devices offer a promising solution to solve this issue. However, harvesting delicate fruits, notably blackberries, poses unique challenges due to their fragility. This study introduces and evaluates a prototype robotic gripper specifically designed for blackberry harvesting. The gripper features an innovative fabric tube mechanism employing motorized twisting action to gently envelop the fruit, ensuring uniform pressure application and minimizing damage. Three types of tubes were developed, varying in elasticity and compressibility using foam padding, spandex, and food-safe cotton cheesecloth. Performance testing focused on assessing each gripper's ability to detach and release blackberries, with emphasis on quantifying damage rates. Results indicate the proposed gripper achieved an 82% success rate in detaching blackberries and a 95% success rate in releasing them, showcasing promised potential for robotic harvesting applications.

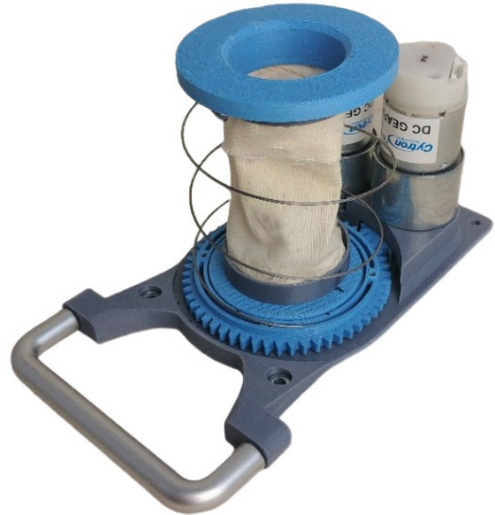

**Fig. 1.** The final developed Berry Twist gripper.

## I. INTRODUCTION

The agricultural industry is grappling with challenges in hiring sufficient workers to meet global food demand, largely due to low wages and adverse health effects [1], [2]. To address this issue, automation, especially in fruit and vegetable harvesting, is being considered as a potential solution. However, there are still hurdles to overcome, particularly in selectively harvesting crops with uneven ripening patterns, and there's a risk of mechanical damage during harvesting, which could affect crop quality [3], [4].

This paper focuses on the robotic harvesting of blackberries, a high-value crop. One of the primary concerns in blackberry harvesting is Red Drupelet Reversion (RDR), where individual drupelets turn reddish after harvesting [5]. According to a survey by Dunteman in 2019, consumers prefer blackberries without RDR, associating it with unripe fruit. The discoloration typically occurs within 24 hours of the fruit entering cool storage [6]. Mechanical injuries resulting from handling during harvesting are identified as a significant cause of RDR, with a recent study reporting that 85% of handled fruit developed RDR compared to 6% of unhandled fruit [5], [7]. Existing harvesting technologies predominantly rely on standard rigid grippers [8], [9]. However, these grippers are not well-suited for handling soft fruits due to their limited adaptability to varying shapes and delicate textures. They exert forces through high and unevenly distributed pressures, posing risks such as bruising, crushing, or deformation [10]. Furthermore, rigid grippers often demand precise positioning and alignment, which makes their application less feasible in the challenging greenhouse environment. In contrast, soft grippers [11], [12], [13], [14] have garnered significant attention in recent years for fruit harvesting due to their capability to handle delicate and easily damaged fruits [15], [16]. By utilizing compliant materials and designs, they offer gentle handling and better conformity to the fruit's shape [17], [18], [19], [20].

Given the issue of RDR, harvesting blackberries requires extra care in distributing forces. In the existing literature, only the recent work by Gunderman et al. [21] focuses on grasping this crop. They introduce a tendon-driven gripper with force feedback but note that the hard support necessary for the force sensor

The work was in part supported under the European Union's Horizon Europe Program from Project EMERGE - Grant Agreement No. 101070918. J.F. Elfferich and D. Dodou are with the BioMechanical Engineering department, Delft University of Technology, Mekelweg 2, 2628 CD Delft, Netherlands d.dodou@tudelft.nl. E. Shahabi and C. Della Santina are with the Cognitive Robotics department, Delft University of Technology, Mekelweg 2, 2628CD Delft, Netherlands. C. Della Santina is with the Institute of Robotics andMechatronics, German Aerospace Center (DLR), 82234 Wesling, GermanyC.DellaSantina@tudelft.nl

caused damage to the berry. In our study, authors extend beyond the already effective pressure distribution properties of fingered soft grippers by proposing a novel grasping modality that applies pressure evenly, ensuring consistent and gentle picking [22]. The proposed gripper, dubbed Berry Twist, consists of a fabric tube - as shown in Figure 1. Two motors twist the fabric from both sides, generating a twisting motion that gently but firmly envelops the fruit. Compared to other non-fingered soft grippers [23], [24], this grasping solution has the added advantage of being inherently food safe. Additionally, the part that comes into contact with the fruit can be easily replaced if damaged or contaminated. The authors conducted a thorough examination of Berry Twist, assessing its performance based on detachment rates, release rates, removal forces, and damage indicators such as leaking drupelets and RDR. Considering the broader perspective, addressing the challenge of blackberry harvesting presents an ideal opportunity for the development of grippers capable of reliably handling an assortment of small and delicate crops. The remainder of the paper is organized as follows. Section II outlines the design and manufacturing process of the mechanism and tubes, along with an explanation of their working principle. In Section III, we present the quantitative results of blackberry grasping, including the percentage of accuracy for grasping and releasing the blackberry. Finally, Section IV concludes the paper and discusses avenues for future research.

## II. Materials And Methods

This study is centered on blackberry harvesting, with a specific goal of safely detaching a single fruit from a plant and relocating it without causing any harm. To achieve this, the gripper must comply with food safety standards outlined in EU regulation No 178/2002. Additionally, we aim for a design that enables pick and place cycles within ten seconds, facilitating efficient coordination with the production process. Moreover, the gripper's design should be adaptable to the dense canopy commonly encountered in most plants, ensuring successful operation across various agricultural environments.

### A. Design and Fabrication

In the prototype (refer to Figure 2), the relatively modest force exerted by the spring ensures that the length of the tube accommodates the stronger twisting motion while maintaining tautness. A single large pre-tensioned spring was positioned externally along the tube's perimeter, eliminating the need for linear guidance and preserving the gripper's deformability. This compliance feature facilitates maneuverability within dense crop canopies, mitigating the risk of crop damage. Additionally, it allows the gripper to adapt its position around the target fruit, especially if it doesn't align precisely with the tube's center. To secure the tube, a custom hose clamp was employed at both ends of the gripper, ensuring stability. The tube boasted an inner diameter of 40 mm, providing ample space for even the largest Blackberry to enter with clearance. Its length was set at 70 mm. For fabrication, all structural components were 3D printed using iglidur® filament (IGUS, I151-PF) for parts in contact with food or sliding over other

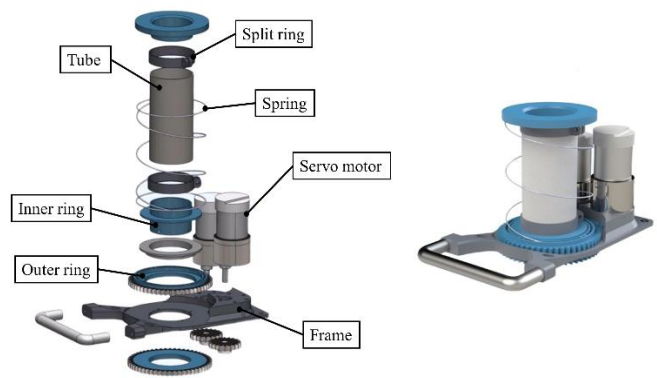

**Fig. 2.** The schematic of the prototype gripper. The left side shows the individual components. On the right side, you'll find the assembled design.

components, guaranteeing food safety. PLA+ filament (eSun) was used for all other components.

The tube itself was crafted from food-safe cloth and sewn into a cylindrical shape using a sewing machine. The spring was custom wound using a 27 mm diameter mandrel and 1 mm thick piano wire, resulting in an inner diameter of 68 mm and a free length of 140 mm. These springs were then slid into U-shaped channels at each end of the gripper and secured in place with setscrews spaced at 90-degree intervals.

As seen in Figure 3, the twisting mechanism in the suggested gripper design rotates a tube's two ends in opposing directions to hold berries. The twisting movement causes the tube's diameter and length to decrease, encasing the fruit in the cloth. All the time the tube stays in touch with the edges of the fruit and wraps around its top and bottom, causing the berry to be completely compressed. This design naturally includes tugging and twisting motions to help break off berries. It is essential to permit flexibility in the spacing between the rings in order to guarantee the grasping mechanism's efficient operation. This distance can be changed by permitting free translation, enabling force-based changes, regulating it with an extra actuator, or linking it with the gripper's rotation. To maintain lightweight and compliant characteristics at the opposite end of the tube, we strategically positioned both servomotors at one end.

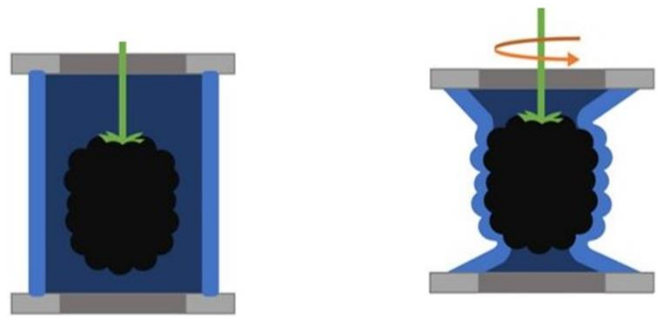

**Fig. 3.** Cross-section demonstrating the twisting tube concept in action, depicting the open configuration (left) and closed configuration (right) during the grasping and detaching of a blackberry. Deformable components are highlighted in blue, the stiff frame in grey, and the actuation is represented by an orange arrow.

In this arrangement, the base motor directly propels the bottom end of the tube, while the other motor rotates the bottom end of the spring. Subsequently, the upper end of the tube effortlessly receives this rotational force. As the gripper tightens, both ends of the tube undergo simultaneous rotation and contraction, commencing from the free end and progressing towards the fixed end. Thus, during gripper closure, the tube twists around the fruit's main axis and exerts a pulling force to detach the fruit from the pedicel. To implement this system, two servos (Cytron, n.d.) with a 120:1 gear ratio were utilized, enabling rotation of the two ends of the twisting tube. Through a modified 55:20 gearbox combination, a rated torque of 1.62 Nm and a stall torque of 6.14 Nm were achieved at the tube's ends. Assuming no losses, the servos operate at a rated rpm of 10.2. To effectively manage the 12V motors, which have a stall current of 1.8 amperes, an appropriate motor shield from DFRobot (n.d.) should be selected.

*B. Tube Materials*

The choice of tube material significantly influences the gripper's overall performance. To thoroughly investigate this aspect, we conducted tests using tubes with varying compressibility, elasticity, and radial elasticity. Seven distinct tubes were manufactured and tested, as illustrated in Figure 4, with detailed characteristics provided in Table I. Our selection process involved opting for cheesecloth as the primary material due to its food-safe properties and non-elastic, woven cotton composition. Additionally, to evaluate the impact of cloth compressibility on performance, we utilized a second cheesecloth variant characterized by its thin and lightweight nature. Moreover, we introduced spandex with a four-way stretch to assess the effects of full elasticity. Given the different surface properties of spandex compared to cheesecloth, we created two additional tubes by sewing thick cheesecloth to the outside and inside of the spandex, enabling an exploration of any performance disparities attributable to surface properties, compressibility, or elasticity.

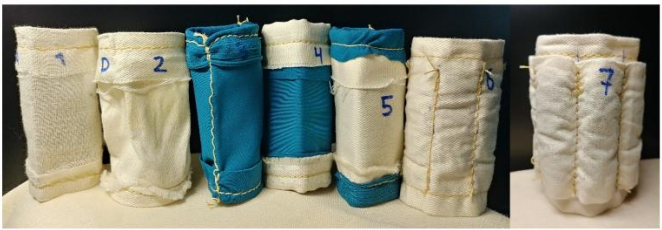

**Fig. 4.** The seven tubes utilized for testing blackberries. Cheesecloth tubes are depicted in white, while spandex tubes are shown in blue. Tube 7, shown on the right, is identical to Tube 6 on the left but reversed inside out.

Furthermore, a cheesecloth tube with internal padding, reversible to provide external padding, was manufactured to introduce radial elasticity. This feature aimed to enable localized deformation for better adaptation to surface irregularities observed in blackberries while leveraging the stiffness of cheesecloth to transmit forces. Radial elasticity was achieved by sewing six strips of 18 mm wide foam along the twisting folds of the tube to facilitate twisting motion. By utilizing the flipped inside-out tube with external padding, we investigated whether differences in gripping performance stemmed from the tube's increased thickness, rendering it stiffer, or its radial elasticity.

TABLE I
Overview of Seven Distinct Tubes Used in the Study

| Tube description | Material |
|---|---|
| Thin cheesecloth | 100% cotton unbleached, 65 grams/m$^2$ |
| Thick Cheesecloth | 100% cotton unbleached, twill weave, 230 grams/m$^2$ |
| Spandex | 82% polyamide ^ 18% elastane, 4-way stretch, 200-225 gram/m$^2$ |
| Thick cheesecloth inside, spandex outside | Cheesecloth, 100% cotton unbleached, twill weave, 230 grams/m$^2$. Spandex, 82% polyamide ^ 18% elastane, 4-way stretch, 200-225 gram/m$^2$ |
| Spandex inside, thick cheesecloth outside | Cheesecloth, 100% cotton unbleached, twill weave, 230 grams/m$^2$. Spandex, 82% polyamide ^ 18% elastane, 4-way stretch, 200-225 gram/m$^2$ |
| Thick cheesecloth, padding inside | Cheesecloth, 100% cotton unbleached, twill weave, 230 grams/m$^2$. Padding, 5mm thick polyether foam SG25 |
| Thick cheesecloth, padding outside | Cheesecloth, 100% cotton unbleached, twill weave, 230 grams/m$^2$. Padding, 5mm thick polyether foam SG25 |

*C. Experimental Setup: Detachment and Removal Measurement*

The purpose of the measurement setup was to assess both the success rate of the gripper and the extent of damage incurred, while also facilitating the measurement of vertical detachment force relative to the pedicel. To achieve this, a custom tensile bench with ample vertical range of motion and equipped with a load cell to measure forces was employed, as depicted in Figure 5. The Twisting-Tube gripper was positioned at the base of the setup and securely fixed to the frame.

In order to accommodate berries of varying pedicel lengths without hindrance, a piece of string was affixed to the end of each pedicel using high-strength tape. This extension allowed the berry to enter the gripper at any desired height, independent of the length of each pedicel. The string was then fastened to a custom clamp directly above the gripper. To maintain consistent vertical positioning of the berry, a custom spacer arm, which could be pivoted downward during berry insertion, was utilized. During testing, this arm could be pivoted away to avoid interference with the gripping process. The submodule of the blackberry, suspended by its pedicel in the clamp, was vertically connected to one end of a load cell. The other end of the load cell was linked to a vertical stage. As such, this load cell could measure the vertical forces acting on the submodule in relation to the vertical stage. These forces included the weight of the submodule and the detachment force exerted by the gripper once it had secured the berry. The vertical stage was powered by a stepper motor, which moved it via a threaded rod. By tracking the pulses sent to the stepper motor and converting them using the pitch of the threaded rod, one could calculate the relative position of the vertical stage.

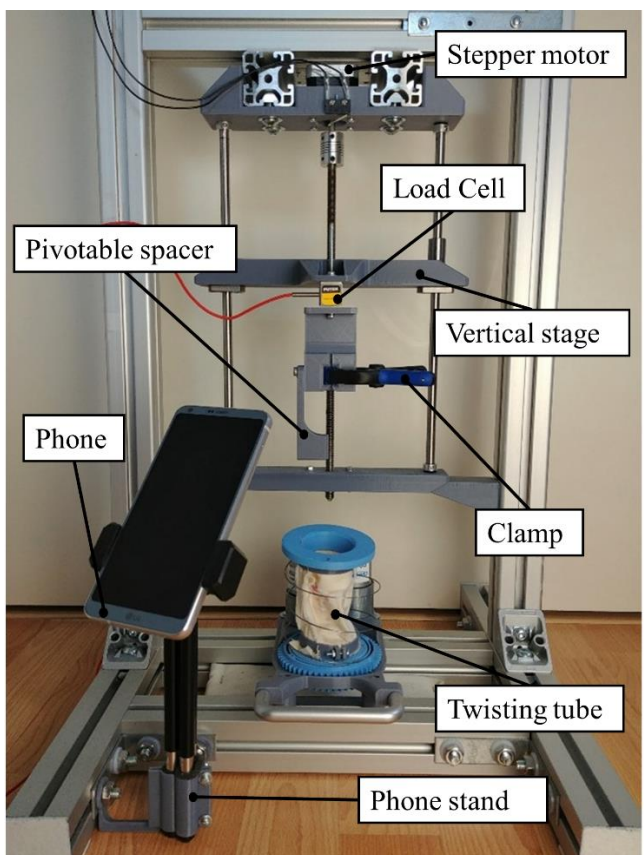

**Fig. 5.** The custom measurement setup.

### III. RESULTS AND DISCUSSION

This section provides the outcomes of the tests conducted using seven different tubes as described above. We start by examining detachment and release rates, followed by pedicel removal forces and rotations of both rings and conclude with an analysis of damage rates in terms of leaking and RDR drupelets. Additionally, authors compare the performance of the Twisting-Tube gripper with different tube materials to understand their impact on its effectiveness.

The blackberries utilized in this study were harvested from a local greenhouse, from De Berkelse Braam, Bosch Fruit B.V. These berries were gathered at a temperature of 19.5°C and with a relative humidity ranging between 70% and 80%. Subsequently, they were transferred to cold storage maintained at 5°C. A total of 18 punnets were filled, each containing 12 blackberries. Among these, 16 punnets contained blackberries with the stem cut approximately 1cm above the sepal, while the remaining two punnets were filled with handpicked berries, serving as a control group. Additionally, a second control group was established by selecting two punnets from the 16 containing blackberries with stems and leaving them unhandled. The remaining 14 punnets with stems, earmarked for testing on the gripper, were divided into two groups of seven for testing on the first and second days, respectively. On each testing day, the seven different types of tubes were evaluated in random order, using blackberries from a single randomly chosen punnet.

During the harvesting process, meticulous care was taken to handle the blackberries gently. Throughout the entire process, no purple stains or signs of leakage from the berries were observed. The blackberries were consistently placed and stored in punnets, ensuring they were never stacked atop one another. Following industry practices, only ripe blackberries, specifically those that were fully black, were selectively picked. All blackberries used in the study were of the Sweet Royalla variety and exhibited an average length of 30.8 mm (SD = 2.8), an average diameter at the widest point of 23.4 mm (SD = 1.9), and an average weight of 7.8 grams (SD = 1.6). These measurements were obtained using a digital caliper on 36 randomly selected blackberries. Research on four different blackberry cultivars has shown similar average sizes and weight ranges.

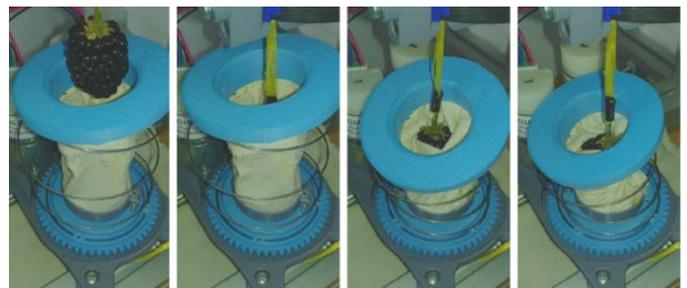

**Fig. 6.** From left to right, the images depict a single pedicel removal process for the Berry Twist gripper equipped with thick cheesecloth. In the initial image, the blackberry enters the gripper. As the process continues, the tube twists and tightens around the tube twists and tightens around the berry. Finally, in the last picture, the pedicel is detached from the berry, completing the procedure.

Table 2 presents the various modes of detachment of the blackberry from the pedicel across the seven different tubes, with an example provided in Figure 6. It's worth noting that in some instances, the pedicel slipped from the tape. The authors opted to exclude these cases as they neither represent a failure nor a success of the gripper. Therefore, the sample size varies for each condition. Additionally, certain potential modes of detachment and failure, such as detachment during the opening of the gripper or wedging the fruit out of the bottom of the gripper, were not observed in any of the tests. The bottom row of Table 2 indicates the percentage of successful detachments relative to the total sample size for each of the seven different conditions. The findings reveal that thick cheesecloth, spandex, and their combinations performed better than other materials, achieving successful detachment in over 75% of the cases. Conversely, other material options performed poorly; for instance, thin cheesecloth often wedged the berry out at the top while tightening, resulting in a successful grip for only 24% of the berries. In most instances, both of the padded tubes also failed to detach the berry, with around a 50% detachment rate.

TABLE II
Variations in Successful and Unsuccessful Detachments of Blackberries from the Pedicel

| | | Thin cheesecloth (n = 21) | Thick cheesecloth (n = 17) | Spandex (n = 16) | Thick cheesecloth inside, spandex outside (n = 17) | Spandex inside, thick cheesecloth outside (n =16) | Thick cheesecloth, padding inside (n = 17) | Thick cheesecloth, padding outside (n = 21) |
|---|---|---|---|---|---|---|---|---|
| Successful detachment at pedicel | During gripping | 5 | 12 | 14 | 12 | 9 | 2 | 8 |
| | During vertical pull | 0 | 2 | 1 | 1 | 4 | 6 | 3 |
| Unsuccessful detachment | No detachment | 2 | 0 | 0 | 0 | 0 | 7 | 10 |
| | Wedging out at top | 14 | 2 | 1 | 3 | 3 | 2 | 0 |
| | Berry ripped apart | 0 | 0 | 0 | 1 | 0 | 0 | 0 |
| | Stem breakage | 0 | 1 | 0 | 0 | 0 | 0 | 0 |
| Gripper successes w.r.t total successful measurements (%) | | 24 | 82 | 94 | 76 | 81 | 47 | 5 |

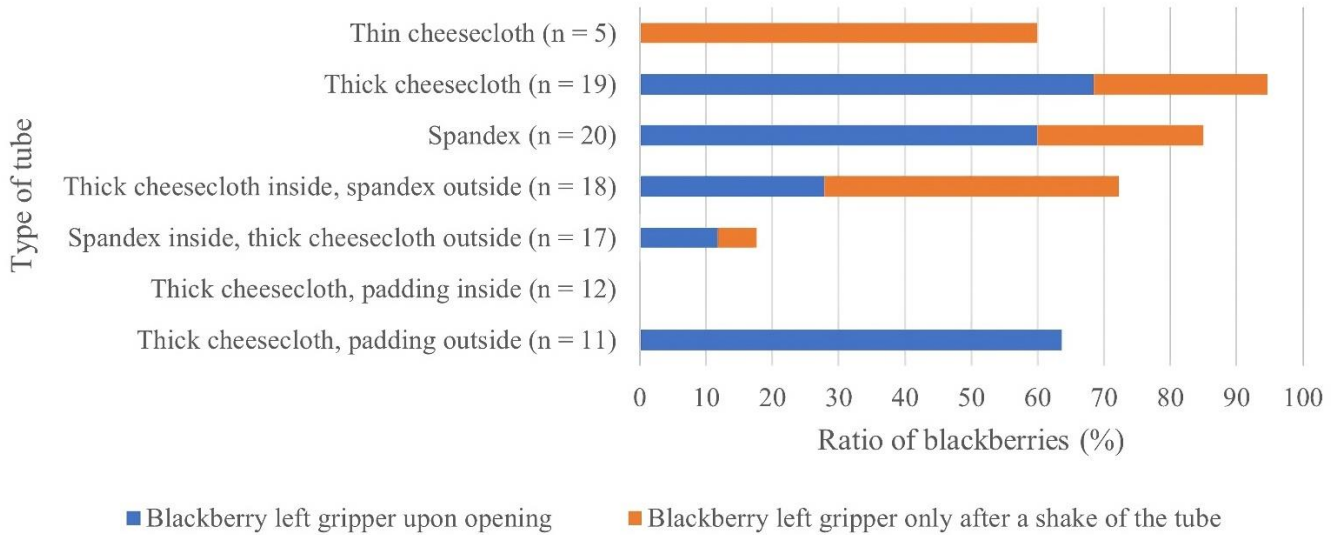

**Fig. 7.** Percentage of successfully released blackberries from the gripper after detachment from the stem during opening or subsequent shaking of the tube.

Another noteworthy observation from the results of Table 2 is that in the majority of cases, the twist and inherent slight pull during the gripping procedure alone were adequate to detach the berry, and only in some instances was the subsequent vertical pull of the vertical stage necessary to achieve successful detachment.

The authors conducted an observation of all the berries that remained in the gripper after the detachment procedure to determine the frequency and timing of berry release (refer to Figure 7). It's important to note that the sample sizes varied, as some berries slipped or wedged out of the gripper during the previous stage and were therefore not present in the gripper for release. Once again, authors observed high success rates for the thick cheesecloth 95% and the spandex 85%. The tube with thick cheesecloth on the inside and spandex on the outside also performed relatively well, releasing 72% of the berries. Conversely, the inverted version exhibited poor performance, with only 18% of the berries successfully released.

## IV. Conclusion

In conclusion, our research presents Berry Twist, a novel twisting-tube gripper tailored for blackberry harvesting, alongside an improved method for assessing Red Drupelet Reversion (RDR) in blackberries. We have demonstrated the gripper's potential, achieving a noteworthy success rate in grasping and releasing blackberries despite employing a predominantly blind grasping strategy. Looking forward, our focus will shift towards developing more intelligent harvesting strategies. We aim to capitalize on sensor technology to enhance the gripper's performance further. Specifically, our future work will involve integrating deformation sensors into the fabric and incorporating a camera at the lower part of the tube. With these advancements, we anticipate unlocking the full potential of Berry Twist and paving the way for more efficient and reliable harvesting solutions for small and delicate crops.